\documentclass[conference]{IEEEtran}
\usepackage{cite}
\usepackage{flushend}
\usepackage{stfloats}
\usepackage{multicol}
\usepackage{graphicx}
\usepackage{epstopdf}
\usepackage{setspace}
\usepackage{lettrine}
\usepackage{amsmath}
\usepackage{amsbsy}
\usepackage{mathrsfs}
\usepackage{amsfonts,amssymb}
\usepackage{algorithm}
\usepackage{algorithmic}
\usepackage{graphicx}
\usepackage{subfigure}
\usepackage{booktabs}
\usepackage{amsmath} 
\usepackage{amssymb}  

\begin{document}

%
\title{ Robustness of Maximum Correntropy\\ Estimation Against Large Outliers}

\author{Badong Chen, \emph{Senior Member, IEEE}, Lei Xing, Haiquan Zhao, \emph{Member, IEEE} \\Bin Xu, \emph{Member, IEEE}, Jos\'e C. Pr\'incipe, \emph{Fellow IEEE} }

\maketitle

\begin{abstract}
The \emph{maximum correntropy criterion} (MCC) has recently been successfully applied in robust regression, classification and adaptive filtering, where the correntropy is maximized instead of minimizing the well-known \emph{mean square error} (MSE) to improve the robustness with respect to outliers (or impulsive noises). Considerable efforts have been devoted to develop various robust adaptive algorithms under MCC, but so far little insight has been gained as to how the optimal solution will be affected by outliers. In this work, we study this problem in the context of parameter estimation for a simple linear \emph{errors-in-variables} (EIV) model where all variables are scalar. Under certain conditions, we derive an upper bound on the absolute value of the estimation error and show that the optimal solution under MCC can be very close to the true value of the unknown parameter even with outliers (whose values can be arbitrarily large) in both input and output variables. An illustrative example is presented to verify and clarify the theory.

\end{abstract}


\let\thefootnote\relax\footnotetext{This work was supported by 973 Program (No. 2015CB351703) and National NSF of China (No. 61372152).
\par Badong Chen, Lei Xing, and Jos\'e C. Pr\'incipe are with the Institute of Artificial Intelligence and Robotics, Xi'an Jiaotong University, Xi'an, 710049, China. (chenbd@mail.xjtu.edu.cn; xl2010@stu.xjtu.edu.cn; principe@cnel.ufl.edu), Jos\'e C. Pr\'incipe is also with the Department of Electrical and Computer Engineering, University of Florida, Gainesville, FL32611 USA.
\par Haiquan Zhao is with the School of Electrical Engineering, Southwest Jiaotong University, Chengdu, China.(hqzhao@home.swjtu.edu.cn)
\par Bin Xu is with the School of Automation, Northwestern Polytechnical
University (NPU), Xi’an, China. (binxu@nwpu.edu.cn)
}


%
\IEEEpeerreviewmaketitle
\section{Introduction}
Second order statistical measures (e.g. MSE, variance, correlation, etc.) are most widely used in machine learning, signal processing and control applications due to their simplicity and efficiency. The learning performances with these measures will, however, deteriorate dramatically when the data contain outliers (which significantly deviate from the bulk of data). Robust statistical measures against outliers (or impulsive noises) are thus of great practical interests, among which the fractional lower order moments (FLOMs) {\cite{shao1993signal,nikias1995signal}}, least absolute deviation (LAD) {\cite{powell1984least,pollard1991asymptotics,peng2003least}} and M-estimation costs {\cite{rousseeuw2005robust,zou2000least,chan2004recursive}} are two typical examples. In particular, recently the \emph{correntropy} as an interesting local similarity measure provides a promising alternative for robust  learning in impulsive noise environments {\cite{liu2007correntropy,principe2010information,chen2013system,chen2012maximum,singh2010loss,he2011robust,he2014half,he2014robust,singh2009using,zhao2011kernel,wu2015kernel,wang2015variable,shi2014convex,chen2015maximum,zhu2016correntropy,chen2014steady,chen2015convergence,wu2015robust}}. Since correntropy is insensitive to large errors (usually caused by some outliers), it can suppress the adverse effects of outliers with large amplitudes. Under the \emph{maximum correntropy criterion} (MCC), the regression (or adaptive filtering) problem can be formulated as maximizing the correntropy between the desired responses and model outputs {\cite{singh2009using,zhao2011kernel,wu2015kernel,wang2015variable,shi2014convex,chen2015maximum,zhu2016correntropy,chen2014steady,chen2015convergence,wu2015robust,chen2016generalized}}.
\par Up to now, many adaptive algorithms (gradient based, fixed-point based, half-quadratic based, etc.) under MCC have been developed to improve the learning performance in presence of outliers{\cite{singh2010loss,he2011robust,he2014half,he2014robust,singh2009using,zhao2011kernel,wu2015kernel,wang2015variable,shi2014convex,chen2015maximum,zhu2016correntropy}}. However, so far little insight has been gained regarding the impact of outliers on the optimal solution under MCC. In the present work, we will attempt to study this problem in order to get a better understanding of the robustness of MCC criterion. To simplify the analysis, we focus on the problem of parameter estimation for a simple linear \emph{errors-in-variables} (EIV) model {\cite{soderstrom2007errors}} in which all variables are scalar. Under certain conditions, we derive an upper bound on the absolute value of the estimation error. Based on the derived results, we may conclude that the optimal estimate under MCC can be very close to the true value of the unknown parameter even in presence of outliers (whose values can be arbitrarily large) in both input and output variables.

\par The rest of the paper is organized as follows. In section II, we describe the problem under consideration. In section III, we derive the main results. In section IV, we present illustrative examples, and in section V we give the conclusion.

\begin{figure}
	\centering
	\includegraphics[width=2.8in,height=2.4in]{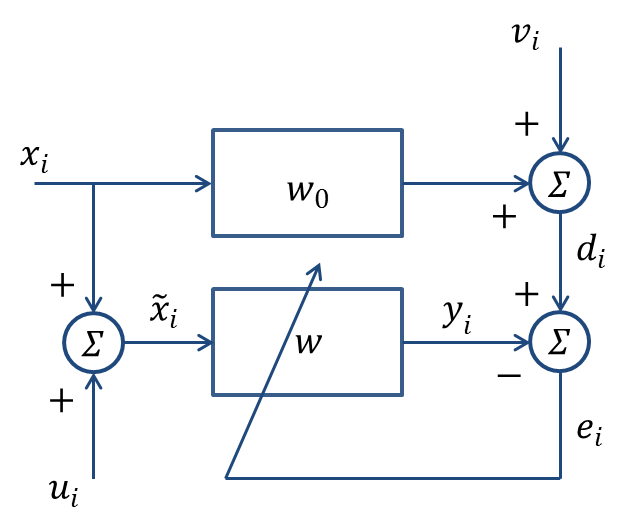}
	\caption{Simple errors-in-variables model}
	\label{fig1}
\end{figure}



\section{MCC BASED PARAMETER ESTIMATION FOR SIMPLE EIV MODEL}

Consider a simple linear EIV model as shown in Fig. 1, where ${w_0} \in \mathbb{R} $ denotes an unknown scalar parameter that needs to be estimated. Let ${x_i}$ be the  true but unobserved input of the unknown system at instant $i$, and ${d_i}$ be the observed output. The observed output and true input of the unknown system are related via
\begin{equation}
{d_i} = {w_0}{x_i} + {v_i}
\end{equation}
where ${v_i}$ denotes the output (observation) noise. In addition, $w$ is the model's parameter, and ${\tilde x_i} = {x_i} + {u_i}$ is the observed input in which ${u_i}$ stands for the input noise. In general, both ${v_i}$ and ${u_i}$ are assumed to be independent of ${x_i}$. The model's output ${y_i}$ is given by
\begin{equation}
{y_i} = w{\tilde x_i} = w({x_i} + {u_i})
\end{equation}
\par Our goal is thus to determine the value of $w$ such that it is as close to ${w_0}$ as possible. A simple approach is to solve $w$ by minimizing the MSE, that is,
\begin{equation}
{w_{MSE}} = \mathop {\arg \min }\limits_{w \in \mathbb{R}} {J_{MSE}}(w) = \mathop {\arg \min }\limits_{w \in \mathbb{R}} \textbf{E}\left[ {e^2} \right]
\end{equation}
where ${e} = {d} - {y}$ is the error between observed output and model output, $\textbf{E}[ \cdot  ]$ denotes the expectation operator, and ${w_{MSE}}$ stands for the optimal solution under MSE. However, the above solution usually leads to inconsistent estimate, i.e. the parameter estimate does not tend to the true value even with very large samples. Some sophisticated methods such as \emph{total least squares} (TLS) {\cite{markovsky2007overview,van1991total,golub1980analysis,markovsky2005application,de1990unifying,roorda1995global}} may give an unbiased estimate, but prior knowledge has to be used and the computational cost is also relatively high.

\par \begin{spacing}{1.0}Another approach is based on the MCC. In this way the model parameter $w$ is determined by {\cite{liu2007correntropy,principe2010information,chen2013system}}
\end{spacing}
\begin{equation}
\begin{aligned}
{w_{MCC}}&= \mathop {\arg \max }\limits_{w \in \mathbb{R}} {J_{MCC}}(w)\\
&= \mathop {\arg \max }\limits_{w \in \mathbb{R}} \textbf{E}\left[ {\exp \left( { - \frac{{e^2}}{{2{\sigma ^2}}}} \right)} \right]
\end{aligned}
\end{equation}
\begin{spacing}{1.0}  \noindent where {${J_{MCC}}(w) =\textbf{E}\left[ {\exp \left( { - \frac{{e^2}}{{2{\sigma ^2}}}} \right)} \right]$ } is the correntropy between ${d}$ and ${y}$ , with $\sigma  > 0$ being the kernel bandwidth, and ${w_{MCC}}$ denotes the corresponding optimal solution. Note that the kernel width $\sigma$ is a key free parameter in MCC, which controls the robustness of the estimator. When the kernel width is very large, the MCC will be approximately equivalent to the MSE criterion. In most practical situations, however, the error distribution is usually unknown, and one has to use the sample mean to approximate the expected value. Given $N$ observed input-output samples $\left\{ {{{\tilde x}_i},{d_i}} \right\}_{i = 1}^N$ , the MCC estimation can be solved by
\end{spacing}
\begin{equation}
\begin{aligned}
{w_{MCC}} &= \mathop {\arg \max }\limits_{w \in \mathbb{R}} {{\hat J}_{MCC}}(w)\\
&{\rm{        }} = \mathop {\arg \max }\limits_{w \in \mathbb{R}} \frac{1}{N}\sum\limits_{i = 1}^N {\exp \left( { - \frac{{{{\left( {{d_i} - w{{\tilde x}_i}} \right)}^2}}}{{2{\sigma ^2}}}} \right)}
\end{aligned}
\end{equation}
where {${\hat J_{MCC}}(w) =\frac{1}{N}\sum\limits_{i = 1}^N {\exp \left( { - \frac{{e_i^2}}{{2{\sigma ^2}}}} \right)}$} is the sample mean estimator of correntropy. Throughout this paper, our notation does not distinguish between random variables and their realizations, which should be clear from the context. It is worth noting that in practical applications, the empirical approximation in (5) is often used as the optimization cost although it does not necessarily approach the expected value when samples go infinite.
\par The non-concave optimization problem in (5) has no closed-form solution but can be effectively solved by using some iterative algorithms such as gradient based methods {\cite{singh2009using,zhao2011kernel}}, fixed-point methods {\cite{chen2015maximum,chen2015convergence}}, half-quadratic methods {\cite{he2011robust,he2014half,he2014robust}}, or evolutionary algorithms such as estimation of distribution algorithm (EDA) {\cite{Chen2007On,Rastegar2005A,Zhang2004On}}.

\section{MAIN RESULTS}
Before proceeding, we give some notations and assumptions. Let
${\varepsilon _u} \ge 0$ and ${\varepsilon _v} \ge 0$ be two non-negative numbers, ${I_N} = \left\{ {1,2, \cdots ,N} \right\}$ be the sample index set, and $I\left( {{\varepsilon _u},{\varepsilon _v}} \right) = \left\{ {i:i \in {I_N},\left| {{u_i}} \right| \le {\varepsilon _u},\left| {{v_i}} \right| \le {\varepsilon _v}} \right\}$ be a subset of ${I_N}$ satisfying $\forall i \in I\left( {{\varepsilon _u},{\varepsilon _v}} \right)$ , $\left| {{u_i}} \right| \le {\varepsilon _u},\left| {{v_i}} \right| \le {\varepsilon _v}$ . In addition, the following two assumptions are made:
\par \emph{Assumption1}: {$N > \left| {I\left( {{\varepsilon _u},{\varepsilon _v}} \right)} \right| = M > \frac{N}{2}$}, where $\left| {I\left( {{\varepsilon _u},{\varepsilon _v}} \right)} \right|$ denotes the cardinality of the set $I\left( {{\varepsilon _u},{\varepsilon _v}} \right)$;
\par \emph{Assumption2}: $\exists c > 0$ such that $\forall i \in I\left( {{\varepsilon _u},{\varepsilon _v}} \right)$ , $\left| {{{\tilde x}_i}} \right| \ge c$ .
\par \begin{spacing}{1.0}\textbf{Remark 1}: The \emph{Assumption 1} means that there are $M$ ( more than $\frac{N}{2}$ ) samples in which the amplitudes of the input and output noises satisfy $\left| {{u_i}} \right| \le {\varepsilon _u},\left| {{v_i}} \right| \le {\varepsilon _v}$, and $N-M$ (at least one) samples that may contain large outliers with $\left| {{u_i}} \right| > {\varepsilon _u}$ or $\left| {{v_i}} \right| > {\varepsilon _v}$ (possibly $\left| {{u_i}} \right| \gg {\varepsilon _u}$ or $\left| {{v_i}} \right| \gg {\varepsilon _v}$ ). The \emph{Assumption 2} is reasonable since for a finite number of samples, the minimum amplitude is in general larger than zero.\end{spacing}

\par With the above notations and assumptions, the following theorem holds:
\par \emph{Theorem 1}: If $\sigma \! > \!\!\frac{{{\varepsilon _v} + \left| {{w_0}} \right|{\varepsilon _u}}}{{\sqrt {2\log \frac{M}{{N - M}}} }}$, then the optimal solution ${w_{MCC}}$ under MCC criterion satisfies $\left| {{w_{MCC}} \!\!-\! {w_0}} \right| \!\le\! \xi $, where\\
\begin{small}$\xi  \!=\! \frac{1}{c}\!\!\left( \!{\sqrt { \!- 2{\sigma ^2}\log \!\left( \!{\exp\! \left( { \!-\! \frac{{{{\left( {{\varepsilon _v} + \left| {{w_0}} \right|{\varepsilon _u}} \right)}^2}}}{{2{\sigma ^2}}}} \right) \!-\! \frac{{N - M}}{M}} \right)}  \!+\! {\varepsilon _v} \!\!+\! \left| {{w_0}} \right|{\varepsilon _u}} \!\right)\!$\end{small}.
\par \emph{Proof}: Since ${w_{MCC}} = \mathop {\arg \max }\limits_{w \in {\rm{\mathbb{R}}}} {\hat J_{MCC}}(w)$ , we have ${\hat J_{MCC}}({w_0}) \le {\hat J_{MCC}}({w_{MCC}})$. To prove $\left| {{w_{MCC}} - {w_0}} \right| \le \xi $, it will suffice to prove ${\hat J_{MCC}}(w) < {\hat J_{MCC}}({w_0})$ for any $w$ satisfying $\left| {w - {w_0}} \right| > \xi $ . Since $N > M > \frac{N}{2}$, we have $0 < \frac{{N - M}}{M} < 1$. As $\sigma  > \frac{{{\varepsilon _v} + \left| {{w_0}} \right|{\varepsilon _u}}}{{\sqrt {2\log \frac{M}{{N - M}}} }}$ , it follows easily that
\begin{equation}
0 < \exp \left( { - \frac{{{{\left( {{\varepsilon _v} + \left| {{w_0}} \right|{\varepsilon _u}} \right)}^2}}}{{2{\sigma ^2}}}} \right) - \frac{{N - M}}{M} < 1
\end{equation}
Further, if {$\left| {w - {w_0}} \right| > \xi $ }, we have $\forall i \in I\left( {{\varepsilon _u},{\varepsilon _v}} \right)$ ,
\begin{equation}
\begin{aligned}
\left| {{e_i}} \right| &= \left| {\left( {{w_0} - w} \right){{\tilde x}_i} + {v_i} - {w_0}{u_i}} \right|\\
&{\rm{    }}\mathop \ge^{(a)} \left| {{w_0} - w} \right| \times \left| {{{\tilde x}_i}} \right| - \left| {{v_i} - {w_0}{u_i}} \right|\\
&{\rm{    }}\mathop  > \limits^{(b)} \xi c - \left( {{\varepsilon _v} + \left| {{w_0}} \right|{\varepsilon _u}} \right)\\
&{\rm{    }} = \sqrt { \!- 2{\sigma ^2}\log\! \left(\! {\exp \left( { - \frac{{{{\left( {{\varepsilon _v} + \left| {{w_0}} \right|{\varepsilon _u}} \right)}^2}}}{{2{\sigma ^2}}}} \right) \!-\! \frac{{N - M}}{M}} \right)}
\end{aligned}
\end{equation}
where (a) comes from $
\left| {\left( {{w_0} - w} \right){{\tilde x}_i} + {v_i} - {w_0}{u_i}} \right| \ge \left| {\left( {{w_0} - w} \right){{\tilde x}_i}} \right| - \left| {{v_i} - {w_0}{u_i}} \right|$ and $\left| {\left( {{w_0} - w} \right){{\tilde x}_i}} \right| = \left| {{w_0} - w} \right|\left| {{{\tilde x}_i}} \right|$, and (b) follows from the \emph{Assumption 2} and $\left| {w \!-\! {w_0}} \right| \!>\! \xi $ and $\left| {{v_i} - {w_0}{u_i}} \right| \le {\varepsilon _v} + \left| {{w_0}} \right|{\varepsilon _u}$ . Thus $\forall i \in I\left( {{\varepsilon _u},{\varepsilon _v}} \right)$,
\begin{equation}
\begin{aligned}
&\exp \left( { - \frac{{e_i^2}}{{2{\sigma ^2}}}} \right) \\
&< \exp \left( { \!-\! \frac{{ \!\!- 2{\sigma ^2}\log \!\!\left( {\exp \!\! \left( { \!- \frac{{{{\left( {{\varepsilon _v} \!+\! \left| {{w_0}} \right|{\varepsilon _u}} \right)}^2}}}{{2{\sigma ^2}}}} \right) \!\!-\! \frac{{N \!-\! M}}{M}} \right)}}{{2{\sigma ^2}}}} \right)\\
&{\rm{                  }} = \exp \left( { - \frac{{{{\left( {{\varepsilon _v} + \left| {{w_0}} \right|{\varepsilon _u}} \right)}^2}}}{{2{\sigma ^2}}}} \right) - \frac{{N - M}}{M}
\end{aligned}
\end{equation}
Then we have ${\hat J_{MCC}}(w) < {\hat J_{MCC}}({w_0})$ for any $w$ satisfying $\left| {w - {w_0}} \right| > \xi $, because
\begin{footnotesize}
	\begin{equation}
	\begin{aligned}
	&{{\hat J}_{MCC}}\left( w \right)\! \\
	&=\! \frac{1}{N}\left\{ {\sum\limits_{i \in I\left( {{\varepsilon _u},{\varepsilon _v}} \right)} {\exp \left( { - \frac{{e_i^2}}{{2{\sigma ^2}}}} \right)}  \!+\!\!\!\! \sum\limits_{i \notin I\left( {{\varepsilon _u},{\varepsilon _v}} \right)}\!\! {\exp \left( { - \frac{{e_i^2}}{{2{\sigma ^2}}}} \right)} } \right\}\\
	{\rm{           }}& <\!\! \frac{1}{N}\!\!\left\{ \!\! {\sum\limits_{i \in I\left( {{\varepsilon _u},{\varepsilon _v}} \right)}\!\!\!\!\! {\left(\!\! {\exp\!\! \left( {\!\! -\! \frac{{{{\left( {{\varepsilon _v} \!+\! \left| {{w_0}} \right|{\varepsilon _u}} \right)}^2}}}{{2{\sigma ^2}}}} \right) \!\!\!-\! \frac{{N \!\!-\! M}}{M}}\!\! \right)} \! +\!\!\!\!\! \sum\limits_{i \notin I\left( {{\varepsilon _u},{\varepsilon _v}} \right)}\!\!\!\!\! {\exp\!\! \left( { \!\!-\! \frac{{e_i^2}}{{2{\sigma ^2}}}} \right)\!\!\!} } \right\}\\
	{\rm{          }}&\mathop  < \limits^{(c)}\!\! \frac{1}{N}\left\{ {\sum\limits_{i \in I\left( {{\varepsilon _u},{\varepsilon _v}} \right)} {\left( {\exp \left( { \!\!-\!\! \frac{{{{\left( {{\varepsilon _v} + \left| {{w_0}} \right|{\varepsilon _u}} \right)}^2}}}{{2{\sigma ^2}}}} \right) \!\!-\!\! \frac{{N \!\!- M}}{M}} \right)} \!\! + N \!\!- M} \right\}\\
	{\rm{          }} &= \frac{1}{N}\sum\limits_{i \in I\left( {{\varepsilon _u},{\varepsilon _v}} \right)} {\exp \left( { - \frac{{{{\left( {{\varepsilon _v} + \left| {{w_0}} \right|{\varepsilon _u}} \right)}^2}}}{{2{\sigma ^2}}}} \right)} \\
	{\rm{          }}&\mathop  \le \limits^{(d)} \frac{1}{N}\sum\limits_{i \in I\left( {{\varepsilon _u},{\varepsilon _v}} \right)} {\exp \left( { - \frac{{{{\left( {{v_i} - {w_0}{u_i}} \right)}^2}}}{{2{\sigma ^2}}}} \right)} \\
	{\rm{          }} &< \frac{1}{N}\sum\limits_{i = 1}^N {\exp \left( { - \frac{{{{\left( {{v_i} - {w_0}{u_i}} \right)}^2}}}{{2{\sigma ^2}}}} \right)}  = {{\hat J}_{MCC}}({w_0})
	\end{aligned}
	\end{equation}
\end{footnotesize}\noindent where (c) comes from $\exp \! \left( { \!-\! \frac{{e_i^2}}{{2{\sigma ^2}}}} \right) \!\le\! 1$, and (d) follows from ${\varepsilon _v} \!+\! \left| {{w_0}} \right|{\varepsilon _u} \!\ge\! \left| {{v_i} \!-\! {w_0}{u_i}} \right|$, $\forall i \!\in\! I\left( {{\varepsilon _u},{\varepsilon _v}} \right)$ . This completes the proof.


\par The following two corollaries are direct consequences of \emph{Theorem 1}.
\par \emph{Corollary 1}: Assume that ${\varepsilon _v} + \left| {{w_0}} \right|{\varepsilon _u} > 0$, and let $\sigma  = \frac{{\lambda \left( {{\varepsilon _v} + \left| {{w_0}} \right|{\varepsilon _u}} \right)}}{{\sqrt {2\log \frac{M}{{N - M}}} }}$ , with $\lambda>1$. Then the optimal solution ${w_{MCC}}$ under MCC satisfies $\left| {{w_{MCC}} - {w_0}} \right| \le \xi $, where
\begin{equation}
\xi  \!=\! \frac{1}{c}\left( {\lambda \sqrt {\frac{{\log \left( {{{\left( {\frac{{N - M}}{M}} \right)}^{{1 \mathord{\left/
									{\vphantom {1 {{\lambda ^2}}}} \right.
									\kern-\nulldelimiterspace} {{\lambda ^2}}}}} \!\!-\! \frac{{N - M}}{M}} \right)}}{{\log \frac{{N - M}}{M}}}}  + 1} \right)\!\!\left( {{\varepsilon _v} \!+\! \left| {{w_0}} \right|{\varepsilon _u}} \right)
\end{equation}
\par \emph{Corollary 2}:  If ${\varepsilon _v} + \left| {{w_0}} \right|{\varepsilon _u} = 0$ , then the optimal solution ${w_{MCC}}$ under MCC satisfies $\left| {{w_{MCC}} - {w_0}} \right| \le \xi $, where
\begin{equation}
\xi  = \frac{\sigma }{c}\sqrt {2\log \left( {\frac{M}{{2M - N}}} \right)}
\end{equation}
\par \textbf{Remark 2}: According to \emph{Corollary 1}, if ${\varepsilon _v} + \left| {{w_0}} \right|{\varepsilon _u} > 0$ and kernel width $\sigma$ is larger than a certain value, the absolute value of the estimation error ${\varepsilon _{MCC}} = {w_{MCC}} - {w_0}$ will be upper bounded by (10). In particular, if both ${\varepsilon _v}$ and ${\varepsilon _u}$ are very small, the upper bound $\xi $ will also be very small. This implies that the MCC solution ${w_{MCC}}$ can be very close to the true value ( $w_0$ ) even in presence of $(N - M)$ outliers whose values can be arbitrarily large, provided that there are $M$( $M > {N \mathord{\left/
		{\vphantom {N 2}} \right.
		\kern-\nulldelimiterspace} 2}$ ) samples disturbed by small noises (bounded by ${\varepsilon _v}$ and ${\varepsilon _u}$ ). In the extreme case, as stated in \emph{Corollary 2}, if ${\varepsilon _v} + \left| {{w_0}} \right|{\varepsilon _u} = 0$ , we have $\xi  \to 0 + $ as $\sigma  \to 0 + $ . In this case, the MCC estimation is almost unbiased as the kernel width $\sigma$ is small enough.
\par  It is worth noting that, due to the inequalities used in the derivation, the real errors in practical situations are usually much smaller and rather far from the derived upper bound $\xi$. This fact will be confirmed by the simulation results provided in the next section.

\textbf{Remark 3}: Although the analysis results in this paper cannot be applied directly to improve the estimation performance in practice, they explain clearly why and how the MCC estimation is robust with respect to outliers especially those with large amplitudes. In addition, according to Theorem 1 and Corollary 1-2, the kernel bandwidth $\sigma$ plays an important role in MCC, which should be set to a proper value (possibly close to the threshold $\frac{{{\varepsilon _v} + \left| {{w_0}} \right|{\varepsilon _u}}}{{\sqrt {2\log \frac{M}{{N - M}}} }}$ ) so as to achieve the best performance. How to optimize the bandwidth $\sigma$ in practice is however a very complicated problem and is left open in this work.

\textbf{Remark 4}: In robust statistics theory, there is a very important concept called breakdown point, which quantifies the smallest proportion of ``bad" data in a sample that a statistics can tolerate before returning arbitrary values. The MCC estimator is essentially a redescending M-estimator, whose breakdown point has been extensively studied in the literature {\cite{huber1984finite,Chen2004On,Yohai1987High,Ricardo1991The,Davis2000Breakdown}}. In particular, it has been shown that the breakdown point of the redescending M-estimators with a bounded objective function can be very close to 1/2 in the location estimation {\cite{huber1984finite,Chen2004On}}. This work however investigates the robustness of a special redescending M-estimator, namely the MCC estimator, in different ways: 1) an EIV model is considered; 2) a bound on the estimation error is derived.

\begin{figure}
	\centering
	\includegraphics[width=\linewidth,height=2.9in]{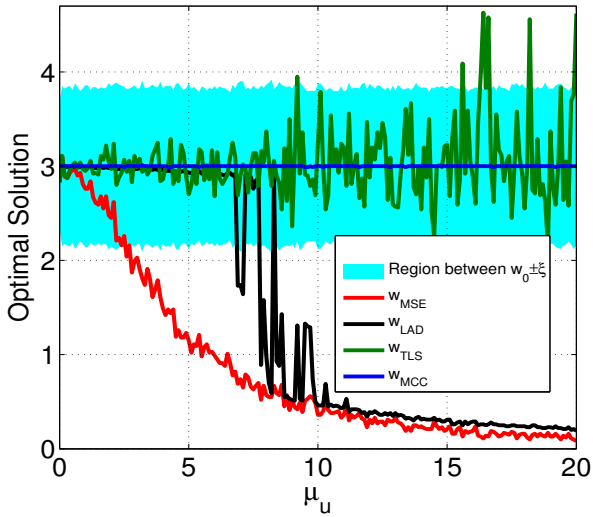}
	\caption{Optimal solutions $w_{MSE}$, $w_{LAD}$, $w_{TLS}$, $w_{MCC}$ and the region between ${w_0} \pm \xi $ with different ${\mu _u}$ ( ${\mu _v}=10.0$ )}
	\label{fig2}
\end{figure}

\begin{figure}
	\centering
	\includegraphics[width=\linewidth,height=2.9in]{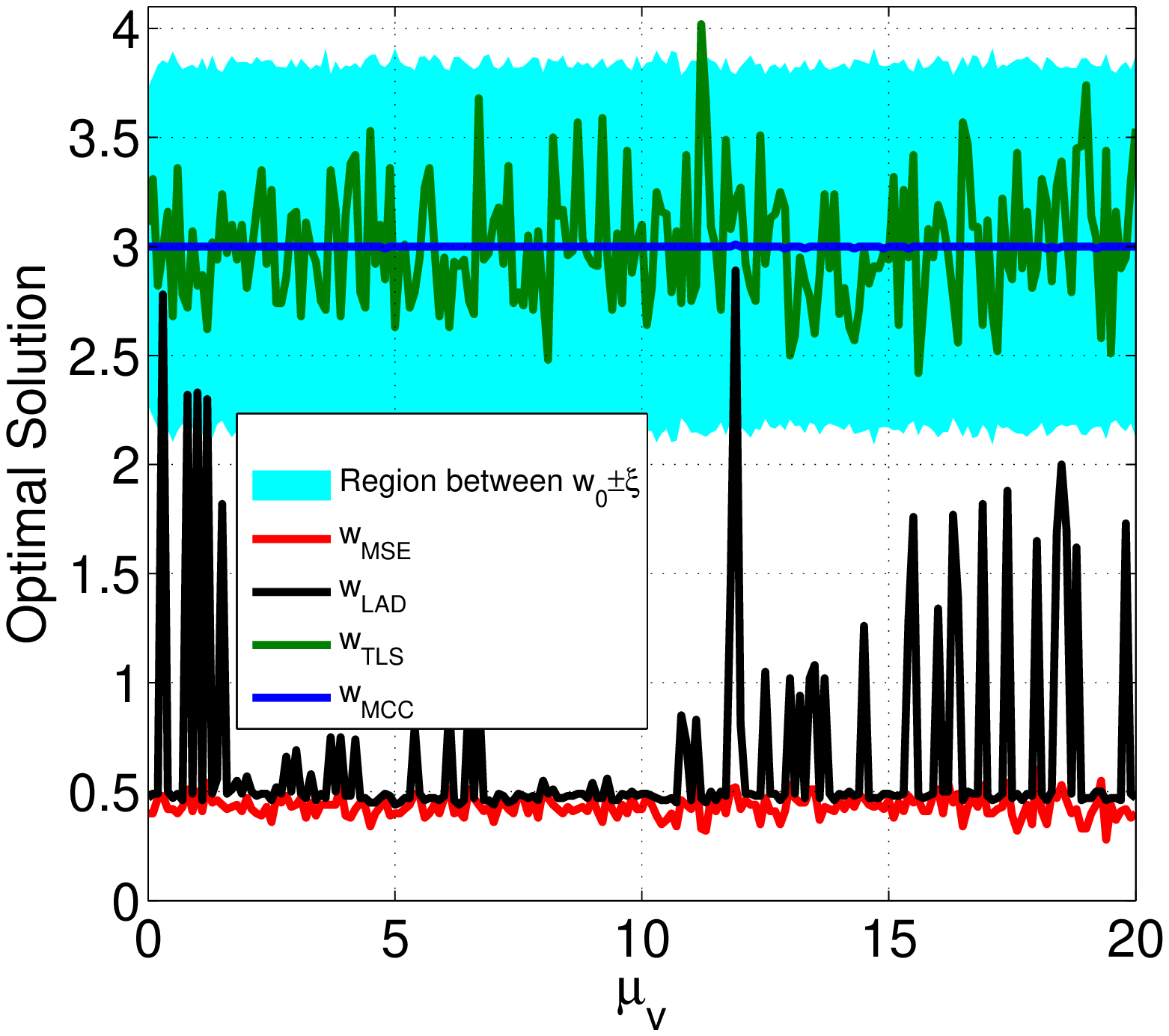}
	\caption{Optimal solutions $w_{MSE}$, $w_{LAD}$, $w_{TLS}$, $w_{MCC}$ and the region between ${w_0} \pm \xi $ with different ${\mu _v}$ ( ${\mu _u}=10.0$ )}
	\label{fig3}
\end{figure}

\begin{table*}[htbp]
	\centering
	\caption{Mean $\pm$ deviation results of the optimal solutions under MSE, LAD, TLS and MCC with different $\mu_u$ ( $\mu_v=10.0$ )}
	\begin{tabular}{ccccc}
		\toprule
		& $w_{MSE}$&$w_{LAD}$ &$w_{TLS}$ & $w_{MCC}$\\
		\midrule
		$\mu_u=0$ & ${\rm{2}}.{\rm{9975}} \pm 0.0{\rm{730}}$& ${\rm{2}}.{\rm{9988}} \pm 0.00{\rm{35}}$ &${\rm{2}}.{\rm{9987}} \pm 0.{\rm{0730}}$ & ${\rm{2}}.{\rm{9996}} \pm 0.{\rm{0020}} $\\
		$\mu_u=2$ & ${\rm{2}}.{\rm{4133}} \pm 0.0{\rm{811}}$& ${\rm{2}}.{\rm{9802}} \pm 0.00{\rm{45}}$ &${\rm{3}}.{\rm{0014}} \pm 0.0{\rm{862}}$ & ${\rm{2}}.{\rm{9996}} \pm 0.{\rm{0020}} $\\
		$\mu_u=4$ & ${\rm{1}}.{\rm{5372}} \pm 0.0{\rm{849}}$& ${\rm{2}}.{\rm{9552}} \pm 0.00{\rm{70}}$ &${\rm{3}}.{\rm{0205}} \pm 0.{\rm{1333}}$ & ${\rm{2}}.{\rm{9996}} \pm 0.{\rm{0024}}$\\
		$\mu_u=6$ & ${\rm{0}}.{\rm{9418}} \pm 0.0{\rm{641}}$& ${\rm{2}}.{\rm{9185}} \pm 0.0{\rm{154}}$ &${\rm{2}}.{\rm{9856}} \pm 0.{\rm{1751}}$ & ${\rm{2}}.{\rm{9998}} \pm 0.{\rm{0014}}$\\
		$\mu_u=8$ & ${\rm{0}}.{\rm{6074}} \pm 0.0{\rm{631}}$& ${\rm{1}}.{\rm{7930}} \pm 0.{\rm{8509}}$ &${\rm{3}}.{\rm{0343}} \pm 0.{\rm{2450}}$ & ${\rm{2}}.{\rm{9991}} \pm 0.{\rm{0029}}$\\
		$\mu_u=10$ & ${\rm{0}}.{\rm{4293}} \pm 0.0{\rm{439}}$& ${\rm{0}}.{\rm{5256}} \pm 0.{\rm{1526}}$ &${\rm{3}}.{\rm{0052}} \pm 0.{\rm{3257}}$ & ${\rm{2}}.{\rm{9993}} \pm 0.{\rm{0026}}$\\
		$\mu_u=12$ & ${\rm{0}}.{\rm{3087}} \pm 0.0{\rm{365}}$& ${\rm{0}}.{\rm{3896}} \pm 0.0{\rm{220}}$ &${\rm{3}}.{\rm{0191}} \pm 0.{\rm{3568}}$ & ${\rm{2}}.{\rm{9994}} \pm 0.{\rm{0024}}$\\
		$\mu_u=14$ & ${\rm{0}}.{\rm{2321}} \pm 0.0{\rm{372}}$& ${\rm{0}}.{\rm{3222}} \pm 0.0{\rm{169}}$ &${\rm{3}}.{\rm{1170}} \pm 0.{\rm{4324}}$ & ${\rm{2}}.{\rm{9993}} \pm 0.{\rm{0026}}$\\
		$\mu_u=16$ & ${\rm{0}}.{\rm{1867}} \pm 0.0{\rm{330}}$& ${\rm{0}}.{\rm{2776}} \pm 0.0{\rm{139}}$ &${\rm{3}}.{\rm{0590}} \pm 0.{\rm{4976}}$ & ${\rm{2}}.{\rm{9993}} \pm 0.{\rm{0029}}$\\
		$\mu_u=18$ & ${\rm{0}}.{\rm{1471}} \pm 0.0{\rm{268}}$& ${\rm{0}}.{\rm{2370}} \pm 0.0{\rm{126}}$ &${\rm{3}}.{\rm{0695}} \pm 0.{\rm{6149}}$ & ${\rm{2}}.{\rm{9992}} \pm 0.{\rm{0027}}$\\
		$\mu_u=20$ & ${\rm{0}}.{\rm{1178}} \pm 0.0{\rm{249}}$& ${\rm{0}}.{\rm{2098}} \pm 0.0{\rm{121}}$ &${\rm{3}}.{\rm{2022}} \pm 0.{\rm{7023}}$ & ${\rm{2}}.{\rm{9995}} \pm 0.{\rm{0022}}$\\
		\bottomrule
	\end{tabular}
	\label{Table1}
\end{table*}

\begin{table*}[htbp]
	\centering
	\caption{Mean $\pm$ deviation results of the optimal solutions under MSE, LAD, TLS and MCC with different $\mu_v$ ( $\mu_u=10.0$ )}
	\begin{tabular}{ccccc}
		\toprule
		& $w_{MSE}$&$w_{LAD}$ &$w_{TLS}$ & $w_{MCC}$\\
		\midrule
		$\mu_v=0$ & ${\rm{0}}.{\rm{4317}} \pm 0.0{\rm{396}}$ & ${\rm{0}}.{\rm{9822}} \pm 0.{\rm{9708}}$ & ${\rm{3}}.{\rm{0387}} \pm 0.{\rm{2405}}$ & ${\rm{2}}.{\rm{9995}} \pm 0.00{\rm{22}}$\\
		$\mu_v=2$ & ${\rm{0}}.{\rm{4228}} \pm 0.0{\rm{320}}$ & ${\rm{0}}.{\rm{6453}} \pm 0.{\rm{3365}}$ & ${\rm{3}}.{\rm{0083}} \pm 0.{\rm{2361}}$ & ${\rm{2}}.{\rm{9993}} \pm 0.00{\rm{26}}$\\
		$\mu_v=4$ & ${\rm{0}}.{\rm{4293}} \pm 0.0{\rm{382}}$ & ${\rm{0}}.{\rm{5325}} \pm 0.{\rm{1020}}$ & ${\rm{3}}.{\rm{0044}} \pm 0.{\rm{2590}}$ & ${\rm{2}}.{\rm{9993}} \pm 0.00{\rm{32}}$\\
		$\mu_v=6$ & ${\rm{0}}.{\rm{4258}} \pm 0.0{\rm{440}}$ & ${\rm{0}}.{\rm{4855}} \pm 0.0{\rm{851}}$ & ${\rm{2}}.{\rm{9953}} \pm 0.{\rm{2555}}$ & ${\rm{2}}.{\rm{9996}} \pm 0.00{\rm{20}}$\\
		$\mu_v=8$ & ${\rm{0}}.{\rm{4246}} \pm 0.0{\rm{457}}$ & ${\rm{0}}.{\rm{5002}} \pm 0.{\rm{1229}}$ & ${\rm{3}}.{\rm{0568}} \pm 0.{\rm{3098}}$ & ${\rm{2}}.{\rm{9997}} \pm 0.00{\rm{22}}$\\
		$\mu_v=10$ & ${\rm{0}}.{\rm{4242}} \pm 0.0{\rm{500}}$ & ${\rm{0}}.{\rm{5463}} \pm 0.{\rm{2752}}$ & ${\rm{3}}.{\rm{0557}} \pm 0.{\rm{3452}}$ & ${\rm{2}}.{\rm{9998}} \pm 0.00{\rm{20}}$\\
		$\mu_v=12$ & ${\rm{0}}.{\rm{4313}} \pm 0.0{\rm{563}}$ & ${\rm{0}}.{\rm{6213}} \pm 0.{\rm{3542}}$ & ${\rm{3}}.{\rm{0111}} \pm 0.{\rm{3330}}$ & ${\rm{2}}.{\rm{9995}} \pm 0.00{\rm{26}}$\\
		$\mu_v=14$ & ${\rm{0}}.{\rm{4199}} \pm 0.0{\rm{526}}$ & ${\rm{0}}.{\rm{5892}} \pm 0.{\rm{2847}}$ & ${\rm{3}}.{\rm{0546}} \pm 0.{\rm{3501}}$ & ${\rm{2}}.{\rm{9996}} \pm 0.00{\rm{24}}$\\
		$\mu_v=16$ & ${\rm{0}}.{\rm{4253}} \pm 0.0{\rm{503}}$ & ${\rm{0}}.{\rm{6499}} \pm 0.{\rm{3604}}$ & ${\rm{3}}.{\rm{0052}} \pm 0.{\rm{2720}}$ & ${\rm{2}}.{\rm{9997}} \pm 0.00{\rm{17}}$\\
		$\mu_v=18$ & ${\rm{0}}.{\rm{4226}} \pm 0.0{\rm{590}}$ & ${\rm{0}}.{\rm{7551}} \pm 0.{\rm{5376}}$ & ${\rm{3}}.{\rm{0612}} \pm 0.{\rm{3824}}$ & ${\rm{2}}.{\rm{9996}} \pm 0.00{\rm{20}}$\\
		$\mu_v=20$ & ${\rm{0}}.{\rm{4355}} \pm 0.0{\rm{690}}$ & ${\rm{0}}.{\rm{8601}} \pm 0.{\rm{6998}}$ & ${\rm{3}}.{\rm{0207}} \pm 0.{\rm{3434}}$ & ${\rm{2}}.{\rm{9996}} \pm 0.00{\rm{24}}$\\
		\bottomrule
	\end{tabular}
	\label{Table2}
\end{table*}

\section{ILLUSTRATIVE EXAMPLES}
\subsection{EXAMPLE 1}
\begin{spacing}{1.0}
	We assume that the true value of the parameter in Fig.1 is ${w_0} = 3.0$ , and the true input signal $x_i$ is uniformly distributed over $\left[ { \!- 2, \!- 1} \right] \cup \left[ {1,2} \right]$ . The input noise $u_i$ and output noise $v_i$ are assumed to be of Gaussian mixture model, given by
\end{spacing}
\begin{equation}
{u_i} \sim \frac{\alpha }{2}\pmb{\mathscr{N}}\left( { - {\mu _u},\sigma _u^2} \right) + \left( {1 - \alpha } \right)\pmb{\mathscr{N}}\left( {0,\sigma _u^2} \right) + \frac{\alpha }{2}\pmb{\mathscr{N}}\left( {{\mu _u},\sigma _u^2} \right)
\end{equation}
\begin{equation}
{v_i} \sim \frac{\beta }{2}\pmb{\mathscr{N}}\left( { - {\mu _v},\sigma _v^2} \right) + \left( {1 - \beta } \right)\pmb{\mathscr{N}}\left( {0,\sigma _v^2} \right) + \frac{\beta }{2}\pmb{\mathscr{N}}\left( {{\mu _v},\sigma _v^2} \right)
\end{equation}
where $\pmb{\mathscr{N}}\left( {\mu ,\sigma^2 } \right)$ denotes a Gaussian density function with mean $\mu $ and variance $\sigma^2 $, $0 \le \alpha ,\beta  \le 1$ are two weighting factors that control the proportions of the outliers (located around $ \pm {\mu _u}$ or $ \pm {\mu _v}$ ) in the observed input and output signals. In the simulations below,  without mentioned otherwise the variances are $\sigma _u^2 = \sigma _v^2 = 0.001$, and the weighting factors are set to $\alpha  = \beta  = 0.15$ . The MCC solutions are solved by using the estimation of distribution algorithm (EDA){\cite{Chen2007On,Rastegar2005A,Zhang2004On}}.

\par First, we illustrate the optimal solutions under MSE, LAD, TLS and MCC with different amplitudes of outliers. Note that the larger the values of ${\mu _u}$ and ${\mu _v}$, the larger the outliers.  Fig. 2 shows the optimal solutions $w_{MSE}$, $w_{LAD}$, $w_{TLS}$, $w_{MCC}$ and the region between ${w_0} \pm \xi $ with different ${\mu _u}$, where ${\mu _v}$ is fixed at ${\mu _v}=10.0$. For different ${\mu _u}$, 1000 i.i.d. samples $\left\{ {{{\tilde x}_i},{d_i}} \right\}_{i = 1}^{1000}$ are generated, and $\xi$ is computed using (10) with ${\varepsilon _u} = {\varepsilon _v} = 0.07$, $c = \mathop {\min }\limits_{i \in I\left( {{\varepsilon _u},{\varepsilon _v}} \right)} \left| {{{\tilde x}_i}} \right|$ , and $\lambda  = 1.2$. Similarly, Fig. 3 shows the optimal solutions $w_{MSE}$, $w_{LAD}$, $w_{TLS}$, $w_{MCC}$ and the region between ${w_0} \pm \xi $ with different ${\mu _v}$ , where ${\mu _u}$ is fixed at ${\mu _u}=10.0$. The corresponding ``\emph{mean $\pm$ deviation}" results of the optimal solutions over 100 Monte Carlos runs are given in Table I and Table II. From these results we can observe: 1) the MCC solution lies within the region between ${w_0} \pm \xi $, being rather close to the true value ${w_0}=3.0$ and very little influenced by both input and output outliers; 2) the estimation error ${\varepsilon _{MCC}} = {w_{MCC}} - {w_0}$ can be much smaller in amplitude than the upper bound $\xi$; 3) MSE, LAD and TLS solutions are sensitive to outliers and can go far beyond the region between ${w_0} \pm \xi $. Especially, the MSE solutions are very sensitive to the input outliers.
\par  Second, we show how the solutions will be affected by the outliers' occurrence probabilities (namely $\alpha$ and $\beta$). Fig. 4 illustrates the optimal solutions under MSE, LAD, TLS and MCC with different $\alpha$, where other parameters are set to $\beta=\alpha$ and $\mu_u=\mu_v=5.0$. As one can see, the MCC solution will be very close to the true value (almost unbiased) when $\alpha$ is smaller than a certain value, although it will get worse dramatically, going far from the true value as $\alpha$ is further increased. The MSE, LAD and TLS solutions, however, will get worse with $\alpha$ increasing, even when $\alpha$ is very small (namely, the outliers are very sparse). Notice that if $\alpha$ is too large,  the Assumption 1 may not hold and the derived upper bound will be inapplicable.
\par Further, we illustrate in Fig. 5 the optimal solutions under MSE and MCC with different kernel widths ( $\sigma$ ), where $\mu_u$ and $\mu_v$ are $\mu_u=\mu_v=5.0$. As expected, the MCC solution will approach the MSE solution as the kernel width is increased. In order to keep the robustness against large outliers, the kernel width in MCC should be set to a relatively small value in general.

\subsection{EXAMPLE 2}
The problem considered in this study is that of estimating a simple EIV model with only one unknown parameter. It is very important to extend the current results to multi-dimensional dynamic case. This is, however, not straightforward since the inequality (a) in (7) does not hold for multi-dimensional case. Here, we present a simulation study for such case and our simulation results suggest that a dynamic EIV model can also be robustly estimated by MCC. Let's consider a 9-taped FIR system with weight vector $ \textbf{\textit{w}}_0=[0.1, 0.2, 0.3, 0.4, 0.5, 0.4, 0.3, 0.2, 0.1] $. The true input signal $ x_i $ is zero-mean Gaussian with variance 1.0 and the distribution models of the input and output noises are assumed to be the same as those in the previous example. In the simulation, the variances are $\sigma _u^2 = \sigma _v^2 = 0.01$, the weighting factors are $ \alpha=\beta=0.3 $  and 2000 i.i.d samples $\left\{ {{{\tilde x}_i},{d_i}} \right\}$ are generated. The squared weight error norm ${\left\| {\textbf{\textit{w}} - {\textbf{\textit{w}}_0}} \right\|^2}$ of MSE, LAD, TLS and MCC with different amplitudes of input and output outliers are presented in Fig.6 ($ \mu_u $ is fixed at 2.0) and Fig.7 ($ \mu_v $ is fixed at 5.0). From Fig.6 and Fig.7, one can see that the squared weight error norm of MCC is very small and little affected by both input and output outliers, while other methods are sensitive to outliers, especially, to input outliers.

\begin{figure}
	\centering
	\includegraphics[width=\linewidth,height=2.9in]{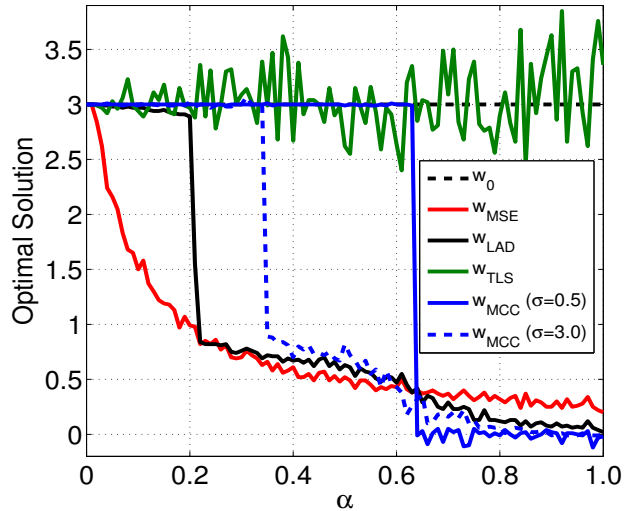}
	\caption{Optimal solutions under MSE, LAD, TLS and MCC with different $\alpha$ ( $\beta=\alpha$, $\mu_u=\mu_v=5.0$)}
	\label{fig4}
\end{figure}

\begin{figure}
	\centering
	\includegraphics[width=\linewidth,height=2.9in]{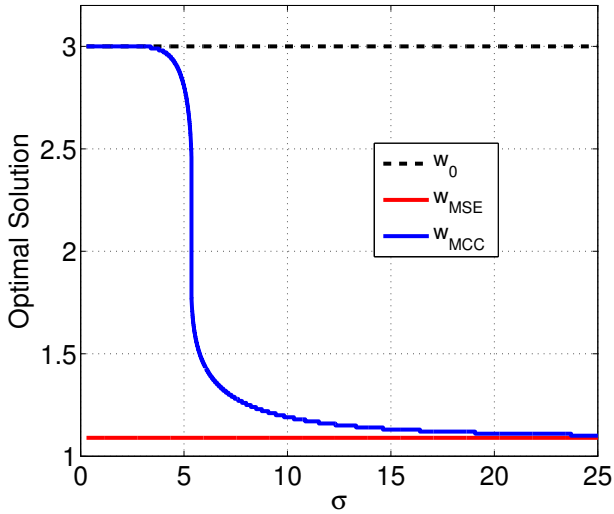}
	\caption{Optimal solutions under MSE and MCC with different kernel widths ( $\mu_u=\mu_v=5.0$ )}
	\label{fig5}
\end{figure}

\begin{figure}
	\centering
	\includegraphics[width=\linewidth,height=2.9in]{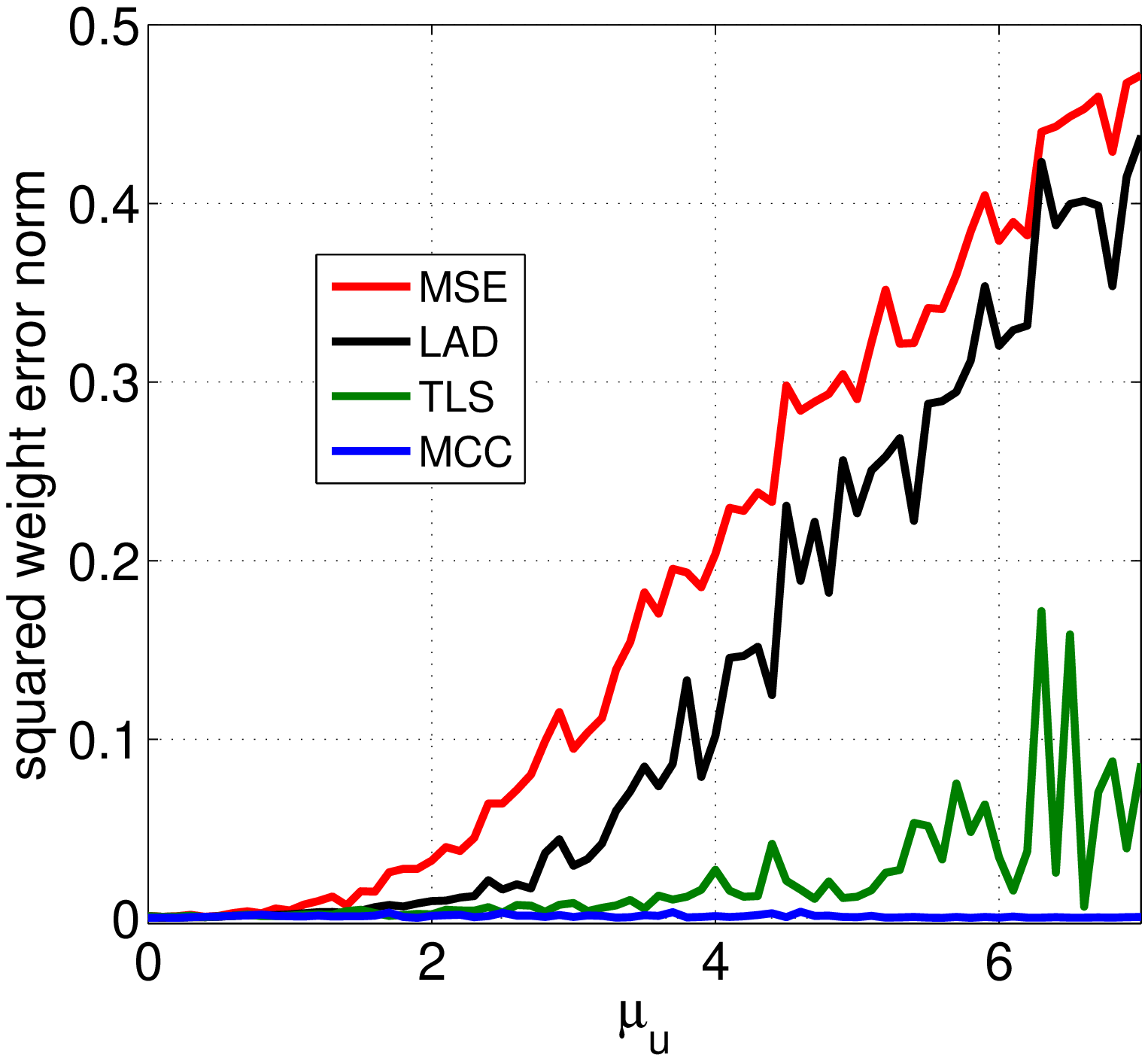}
	\caption{Squared weight error norm of MSE, LAD, TLS and MCC with different ${\mu _u}$ ( ${\mu _v}=2.0$ )}
	\label{fig6}
\end{figure}

\begin{figure}
	\centering
	\includegraphics[width=\linewidth,height=2.9in]{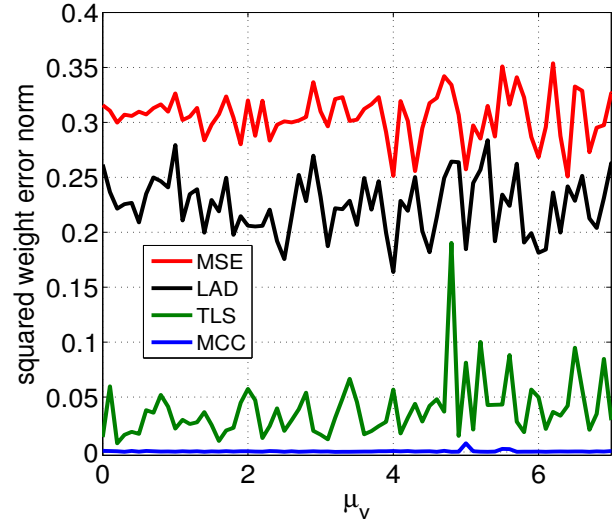}
	\caption{Squared weight error norm of MSE, LAD, TLS and MCC with different ${\mu _v}$ ( ${\mu _u}=5.0$ )}
	\label{fig7}
\end{figure}

\section{CONCLUSION}
\par We investigated in this work the robustness of the maximum correntropy criterion against large outliers, in the context of parameter estimation for a simple linear \emph{errors-in-variables} (EIV) model where all variables are scalar. Under certain conditions, we derived an upper bound on the amplitude of the estimation error. The obtained results suggest that the MCC estimation can be very close to the true value of the unknown parameter even with outliers (whose values can be arbitrarily large) in both input and output variables. The analysis results have been verified by illustrative examples. Extending the results of this study from simple EIV model to multivariable case is however not straightforward. This remains a challenge for future study.

\bibliographystyle{unsrt}

\begin{thebibliography}{10}
	
	\bibitem{shao1993signal}
	Min Shao and Chrysostomos~L Nikias.
	\newblock Signal processing with fractional lower order moments: stable
	processes and their applications.
	\newblock {\em Proceedings of the IEEE}, 81(7):986--1010, 1993.
	
	\bibitem{nikias1995signal}
	Chrysostomos~L Nikias and Min Shao.
	\newblock {\em Signal Processing with Alpha-stable Distributions and
		Applications}.
	\newblock Wiley-Interscience, 1995.
	
	\bibitem{powell1984least}
	James~L Powell.
	\newblock Least absolute deviations estimation for the censored regression
	model ?.
	\newblock {\em Journal of Econometrics}, 25(3):303--325, 1984.
	
	\bibitem{pollard1991asymptotics}
	David Pollard.
	\newblock Asymptotics for least absolute deviation regression estimators.
	\newblock {\em Econometric Theory}, 7(2):186--199, 1991.
	
	\bibitem{peng2003least}
	Liang Peng and Qiwei Yao.
	\newblock Least absolute deviations estimation for arch and garch models.
	\newblock {\em Biometrika}, 90(4):967--975, 2003.
	
	\bibitem{rousseeuw2005robust}
	Peter~J Rousseeuw and Annick~M Leroy.
	\newblock {\em Robust Regression and Outlier Detection}, volume 589.
	\newblock John Wiley \& Sons, 2005.
	
	\bibitem{zou2000least}
	Yuexian Zou, Shing-Chow Chan, and Tung-Sang Ng.
	\newblock Least mean m-estimate algorithms for robust adaptive filtering in
	impulse noise.
	\newblock {\em Circuits and Systems II: Analog and Digital Signal Processing,
		IEEE Transactions on}, 47(12):1564--1569, 2000.
	
	\bibitem{chan2004recursive}
	Shing-Chow Chan and Yue-Xian Zou.
	\newblock A recursive least m-estimate algorithm for robust adaptive filtering
	in impulsive noise: fast algorithm and convergence performance analysis.
	\newblock {\em Signal Processing, IEEE Transactions on}, 52(4):975--991, 2004.
	
	\bibitem{liu2007correntropy}
	Weifeng Liu, Puskal~P Pokharel, and Jos{\'e}~C Pr{\'\i}ncipe.
	\newblock Correntropy: properties and applications in non-gaussian signal
	processing.
	\newblock {\em Signal Processing, IEEE Transactions on}, 55(11):5286--5298,
	2007.
	
	\bibitem{principe2010information}
	Jose~C Principe.
	\newblock {\em Information Theoretic Learning: Renyi's Entropy and Kernel
		Perspectives}.
	\newblock Springer Science \& Business Media, 2010.
	
	\bibitem{chen2013system}
	Badong Chen, Yu~Zhu, Jinchun Hu, and Jose~C Principe.
	\newblock {\em System Parameter Identification: Information Criteria and
		Algorithms}.
	\newblock Newnes, 2013.
	
	\bibitem{chen2012maximum}
	Badong Chen and Jos{\'e}~C Pr{\'\i}ncipe.
	\newblock Maximum correntropy estimation is a smoothed map estimation.
	\newblock {\em Signal Processing Letters, IEEE}, 19(8):491--494, 2012.
	
	\bibitem{singh2010loss}
	Abhishek Singh and Jose~C Principe.
	\newblock A loss function for classification based on a robust similarity
	metric.
	\newblock In {\em Neural Networks (IJCNN), The 2010 International Joint
		Conference on}, pages 1--6. IEEE, 2010.
	
	\bibitem{he2011robust}
	Ran He, Bao-Gang Hu, Wei-Shi Zheng, and Xiang-Wei Kong.
	\newblock Robust principal component analysis based on maximum correntropy
	criterion.
	\newblock {\em Image Processing, IEEE Transactions on}, 20(6):1485--1494, 2011.
	
	\bibitem{he2014half}
	Ran He, Wei-Shi Zheng, Tieniu Tan, and Zhenan Sun.
	\newblock Half-quadratic-based iterative minimization for robust sparse
	representation.
	\newblock {\em Pattern Analysis and Machine Intelligence, IEEE Transactions
		on}, 36(2):261--275, 2014.
	
	\bibitem{he2014robust}
	Ran He, Tieniu Tan, and Liang Wang.
	\newblock Robust recovery of corrupted low-rankmatrix by implicit regularizers.
	\newblock {\em Pattern Analysis and Machine Intelligence, IEEE Transactions
		on}, 36(4):770--783, 2014.
	
	\bibitem{singh2009using}
	Abhishek Singh and Jose~C Principe.
	\newblock Using correntropy as a cost function in linear adaptive filters.
	\newblock In {\em Neural Networks, 2009. IJCNN 2009. International Joint
		Conference on}, pages 2950--2955. IEEE, 2009.
	
	\bibitem{zhao2011kernel}
	Songlin Zhao, Badong Chen, and Jose~C Principe.
	\newblock Kernel adaptive filtering with maximum correntropy criterion.
	\newblock In {\em Neural Networks (IJCNN), The 2011 International Joint
		Conference on}, pages 2012--2017. IEEE, 2011.
	
	\bibitem{wu2015kernel}
	Zongze Wu, Jiahao Shi, Xie Zhang, Wentao Ma, and Badong Chen.
	\newblock Kernel recursive maximum correntropy.
	\newblock {\em Signal Processing}, 117:11--26, 2015.
	
	\bibitem{wang2015variable}
	Ren Wang, Badong Chen, Nanning Zheng, and Jose~C Principe.
	\newblock A variable step-size adaptive algorithm under maximum correntropy
	criterion.
	\newblock In {\em Neural Networks (IJCNN), 2015 International Joint Conference
		on}, pages 1--5. IEEE, 2015.
	
	\bibitem{shi2014convex}
	Liming Shi and Yun Lin.
	\newblock Convex combination of adaptive filters under the maximum correntropy
	criterion in impulsive interference.
	\newblock {\em Signal Processing Letters, IEEE}, 21(11):1385--1388, 2014.
	
	\bibitem{chen2015maximum}
	Badong Chen, Xi~Liu, Haiquan Zhao, and Jos{\'e}~C Pr{\'\i}ncipe.
	\newblock Maximum correntropy kalman filter.
	\newblock {\em arXiv preprint arXiv:1509.04580}, 2015.
	
	\bibitem{zhu2016correntropy}
	Fei Zhu, Abderrahim Halimi, Paul Honeine, Badong Chen, and Nanning Zheng.
	\newblock Correntropy maximization via admm: Application to robust
	hyperspectral unmixing.
	\newblock {\em IEEE Transactions on Geoscience \& Remote Sensing},
	PP(99):1--12, 2016.
	
	\bibitem{chen2014steady}
	Badong Chen, Lei Xing, Junli Liang, Nanning Zheng, and Jos{\'e}~C
	Pr{\'\i}ncipe.
	\newblock Steady-state mean-square error analysis for adaptive filtering under
	the maximum correntropy criterion.
	\newblock {\em Signal Processing Letters, IEEE}, 21(7):880--884, 2014.
	
	\bibitem{chen2015convergence}
	Badong Chen, Jianji Wang, Haiquan Zhao, Nanning Zheng, and Jos{\'e}~C
	Pr{\'\i}ncipe.
	\newblock Convergence of a fixed-point algorithm under maximum correntropy
	criterion.
	\newblock {\em Signal Processing Letters, IEEE}, 22(10):1723--1727, 2015.
	
	\bibitem{wu2015robust}
	Zongze Wu, Siyuan Peng, Badong Chen, and Haiquan Zhao.
	\newblock Robust hammerstein adaptive filtering under maximum correntropy
	criterion.
	\newblock {\em Entropy}, 17(10):7149--7166, 2015.
	
	\bibitem{chen2016generalized}
	Badong Chen, Lei Xing, Haiquan Zhao, Nanning Zheng, and Jos{\'e}~C
	Pr{\'\i}ncipe.
	\newblock Generalized correntropy for robust adaptive filtering.
	\newblock {\em Signal Processing, IEEE Transactions on}, 64(13):3376--3387,
	2016.
	
	\bibitem{soderstrom2007errors}
	Torsten S{\"o}derstr{\"o}m.
	\newblock Errors-in-variables methods in system identification.
	\newblock {\em Automatica}, 43(6):939--958, 2007.
	
	\bibitem{markovsky2007overview}
	Ivan Markovsky and Sabine Van~Huffel.
	\newblock Overview of total least-squares methods.
	\newblock {\em Signal processing}, 87(10):2283--2302, 2007.
	
	\bibitem{van1991total}
	Sabine Van~Huffel and Joos Vandewalle.
	\newblock {\em The Total Least Squares Problem: Computational Aspects and
		Analysis}, volume~9.
	\newblock Siam, 1991.
	
	\bibitem{golub1980analysis}
	Gene~H Golub and Charles~F Van~Loan.
	\newblock An analysis of the total least squares problem.
	\newblock {\em SIAM Journal on Numerical Analysis}, 17(6):883--893, 1980.
	
	\bibitem{markovsky2005application}
	Ivan Markovsky, Jan~C Willems, Sabine Van~Huffel, Bart De~Moor, and Rik
	Pintelon.
	\newblock Application of structured total least squares for system
	identification and model reduction.
	\newblock {\em Automatic Control, IEEE Transactions on}, 50(10):1490--1500,
	2005.
	
	\bibitem{de1990unifying}
	Bart De~Moor and Joos Vandewalle.
	\newblock A unifying theorem for linear and total linear least squares.
	\newblock {\em Automatic Control, IEEE Transactions on}, 35(5):563--566, 1990.
	
	\bibitem{roorda1995global}
	Berend Roorda and Christiaan Heij.
	\newblock Global total least squares modeling of multivariable time series.
	\newblock {\em Automatic Control, IEEE Transactions on}, 40(1):50--63, 1995.
	
	\bibitem{Chen2007On}
	Tianshi Chen, Ke~Tang, Guoliang Chen, and Xin Yao.
	\newblock On the analysis of average time complexity of estimation of
	distribution algorithms.
	\newblock In {\em IEEE Congress on Evolutionary Computation}, pages 453--460,
	2007.
	
	\bibitem{Rastegar2005A}
	R.~Rastegar and M.~R. Meybodi.
	\newblock A study on the global convergence time complexity of estimation of
	distribution algorithms.
	\newblock In {\em International Workshop on Rough Sets, Fuzzy Sets, Data
		Mining, and Granular-Soft Computing}, pages 441--450, 2005.
	
	\bibitem{Zhang2004On}
	Qingfu Zhang and H~Muhlenbein.
	\newblock On the convergence of a class of estimation of distribution
	algorithms.
	\newblock {\em Evolutionary Computation IEEE Transactions on}, 8(2):127--136,
	2004.
	
	\bibitem{huber1984finite}
	Peter~J. Huber.
	\newblock Finite sample breakdown of m- and p-estimators.
	\newblock {\em Annals of Statistics}, 12(1):119--126, 1984.
	
	\bibitem{Chen2004On}
	Zhiqiang Chen and David~E. Tyler.
	\newblock On the finite sample breakdown points of redescending m -estimates of
	location.
	\newblock {\em Statistics \& Probability Letters}, 69(3):233--242, 2004.
	
	\bibitem{Yohai1987High}
	Victor~J. Yohai.
	\newblock High breakdown-point and high efficiency robust estimates for
	regression.
	\newblock {\em Annals of Statistics}, 15(2):642--656, 1987.
	
	\bibitem{Ricardo1991The}
	Ricardo~A. Maronna and Victor~J. Yohai.
	\newblock The breakdown point of simultaneous general m estimates of regression
	and scale.
	\newblock {\em Journal of the American Statistical Association},
	86(415):699--703, 1991.
	
	\bibitem{Davis2000Breakdown}
	R.~A. Davis, Wtm Dunsmuir, and Y.~Wang.
	\newblock Breakdown points of t-type regression estimators.
	\newblock {\em Biometrika}, 87(3):675--687, 2000.
	
\end{thebibliography}

\end{document}